\documentclass[runningheads]{llncs}
\usepackage{graphicx}
\usepackage{comment}
\usepackage{amsmath,amssymb} % define this before the line numbering.
\usepackage{color}
\usepackage{hyperref}
\usepackage{multirow}
\usepackage{booktabs}
\usepackage{tabularx}
\usepackage{bm}
\usepackage{pifont}% http://ctan.org/pkg/pifont
\hypersetup{
	colorlinks=true,
	linkcolor=black,   
	urlcolor=magenta,
}

\newcommand{\bR}{\mathbf}
\newcommand{\cmark}{\ding{51}}%
\newcommand{\xmark}{\ding{55}}%

\begin{document}
% \linenumbers
\pagestyle{headings}
\mainmatter
\def\ECCVSubNumber{2393}  % Insert your submission number here
\title{GSNet: Joint Vehicle Pose and Shape Reconstruction with Geometrical and Scene-aware Supervision} %

% INITIAL SUBMISSION 
\begin{comment}
\titlerunning{ECCV-20 submission ID \ECCVSubNumber} 
\authorrunning{ECCV-20 submission ID \ECCVSubNumber} 
\author{Anonymous ECCV submission}
\institute{Paper ID \ECCVSubNumber}
\end{comment}
%******************

% CAMERA READY SUBMISSION
%\begin{comment}
\titlerunning{GSNet: Joint Vehicle Pose and Shape Reconstruction}
% If the paper title is too long for the running head, you can set
% an abbreviated paper title here
%
\author{Lei Ke\inst{1} \and
Shichao Li\inst{1} \and
Yanan Sun\inst{1} \and 
Yu-Wing Tai\inst{1,2} \and 
Chi-Keung Tang\inst{1}}
\authorrunning{L. Ke, S. Li, Y. Sun, Y.-W. Tai, and C.-K. Tang}
% First names are abbreviated in the running head.
% If there are more than two authors, 'et al.' is used.
%
\institute{The Hong Kong University of Science and Technology \and
Kwai Inc.\\
\email{\{lkeab,slicd,ysuncd,yuwing,cktang\}@cse.ust.hk}}
%\end{comment}
%******************
\maketitle

%%%%%%%%% ABSTRACT
\begin{abstract}	
	We present a novel end-to-end framework named as GSNet (\textbf{\underline{G}}eometric and \textbf{\underline{S}}cene-aware \underline{\textbf{Net}}work), which jointly estimates 6DoF poses and reconstructs detailed 3D car shapes from single urban street view. GSNet utilizes a unique four-way feature extraction and fusion scheme and directly regresses 6DoF poses and shapes in a single forward pass. Extensive experiments show that our diverse feature extraction and fusion scheme can greatly improve model performance. Based on a divide-and-conquer 3D shape representation strategy, GSNet reconstructs 3D vehicle shape with great detail (1352 vertices and 2700 faces). This dense mesh representation further leads us to consider geometrical consistency and scene context, and inspires a new multi-objective loss function to regularize network training, which in turn improves the accuracy of 6D pose estimation and validates the merit of jointly performing both tasks. We evaluate GSNet on the largest multi-task ApolloCar3D benchmark and achieve state-of-the-art performance both quantitatively and qualitatively. Project page is available at \url{https://lkeab.github.io/gsnet/}.

\end{abstract}

%%%%%%%%% BODY TEXT
%------------------------------------------------------------------------
\section{Introduction}
\begin{figure}[!t]
	\centering
	%\fbox{\rule{0pt}{2in} \rule{0.9\linewidth}{0pt}}
	\includegraphics[width=0.8\linewidth]{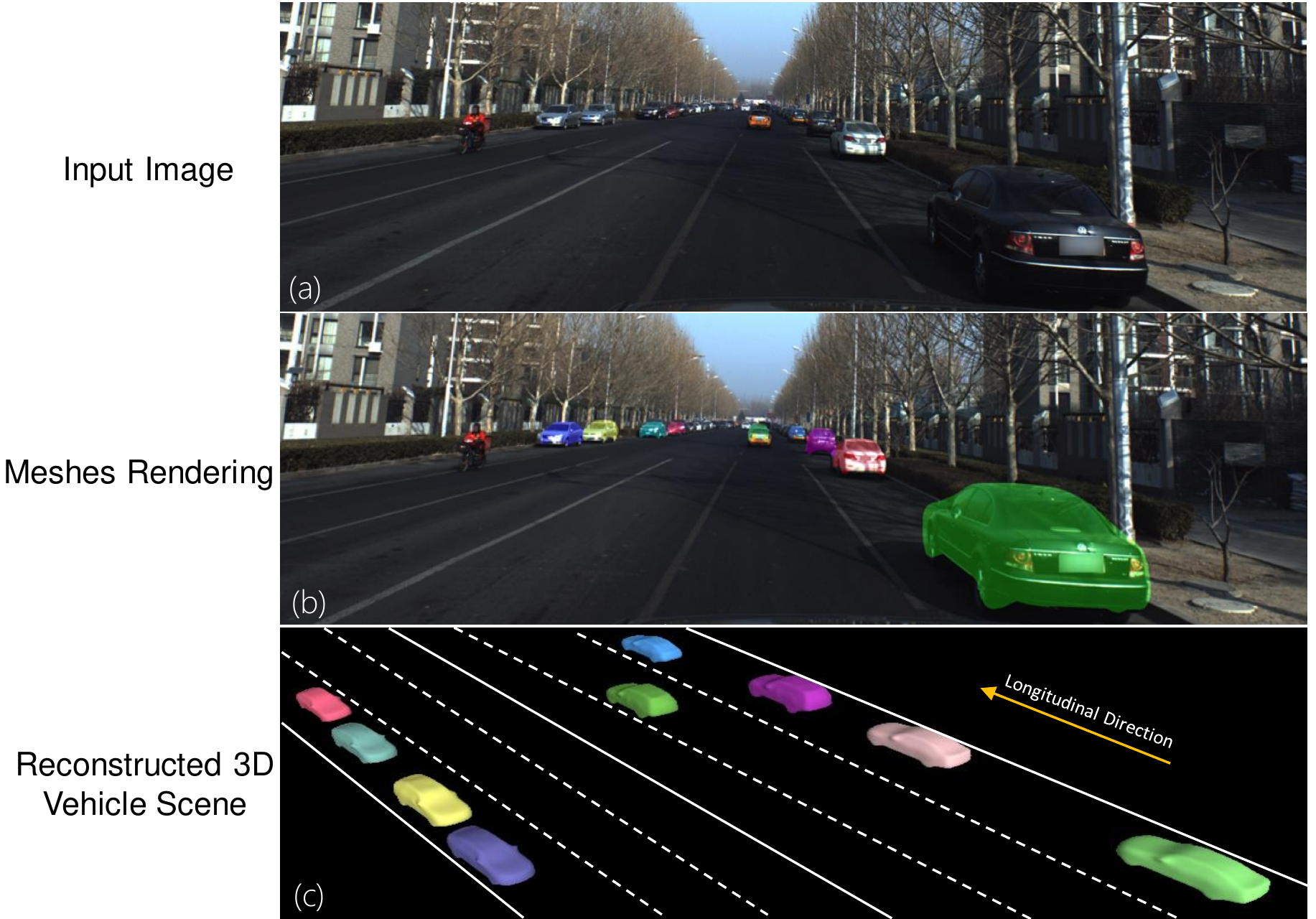}
	\caption{Joint vehicle pose and shape reconstruction results of our GSNet, where (a) is the input RGB image, (b) shows the reconstructed 3D car meshes projected onto the original image, (c) is a novel aerial view of the reconstructed 3D traffic scene. Corresponding car instances in (b) and (c) are depicted in the same color.}
	%\label{fig:model}
	\label{fig:example1}
\end{figure}

Traffic scene understanding is an active area in autonomous driving, where one emerging and challenging task is to perceive 3D attributes (including translation, rotation and shape) of vehicle instances in a dynamic environment as Figure~\ref{fig:example1} shows. Compared to other scene representations such as 2D/3D bounding boxes~\cite{chen2016monocular,li2019stereo,mousavian20173d}, semantic masks~\cite{cordts2016cityscapes,pohlen2017full} and depth maps~\cite{xu2018multi}, representing traffic scene with 6D object pose and detailed 3D shape is more informative for spatial reasoning and motion planning of self-driving cars.

Due to the lack of depth information in monocular RGB images, many existing works resort to stereo camera rigs~\cite{li2019stereo,li2018stereo} or expensive LiDAR~\cite{yang2018pixor,yang2019std,ku2019monocular}.
However, they are limited by constrained perception range~\cite{li2019stereo} or sparse 3D points for distant regions in the front view~\cite{song2019apollocar3d}. When using only a single RGB image, works that jointly reconstruct vehicle pose and shape can be classified into two categories:~\textit{fitting-based} and direct \textit{regression-based}.
~\textit{Fitting-based} methods~\cite{chabot2017deep,su2015render,song2019apollocar3d} use a two-stage strategy where they first extract 2D image cues such as bounding boxes and keypoints and then fit a 3D template vehicle to best match its 2D image observations. The second stage is a post-processing step that is usually time-consuming due to iterative non-linear optimization, making it less applicable for real-time autonomous driving. On the contrary, \textit{regression-based} methods~\cite{kundu20183d,song2019apollocar3d} directly predict 3D pose/shape parameters with a single efficient forward pass of a deep network and is gaining increasing popularity with the growing scale of autonomous driving datasets.

Despite the recent regression-based methods having achieved remarkable performance for joint vehicle pose estimation and 3D shape reconstruction, we point out some unexplored yet valuable research questions: 
(1) Most regression-based networks~\cite{kundu20183d,song2019apollocar3d,li2017deep} inherit classical 2D object detection architectures that solely use region of interest (ROI) features to regress 3D parameters. \textit{How other potential feature representation can improve network performance} is less studied. 
(2) Deep networks require huge amounts of supervision~\cite{kolotouros2019learning}, where useful supervisory signals other than manually annotated input-target pairs are favorable. Consistency brought by projective geometry is one possibility, yet the optimal design is still under-explored. Render-and-compare loss was used in~\cite{kundu20183d} but it suffers from ambiguities where similar 2D projected masks can correspond to different 3D unknown parameters. For example, a mask similar to the ground truth mask is produced after changing the ground truth 3D pose by 180 degrees around the symmetry axis, i.e., the prediction is not penalized enough despite being incorrect.
(3) Previous regression-based works only penalize prediction error for single car instance and separate it from its environmental context, but a traffic scene includes the interaction between multiple instances and the relationship between instances with the physical world. We argue that considering these extra information can improve the training of a deep network.

We investigate these above questions and propose GSNet (\textbf{\underline{G}}eometric and \textbf{\underline{S}}cene-aware \underline{\textbf{Net}}work), an end-to-end multi-task network that can estimate 6DoF car pose and reconstruct dense 3D shape simultaneously. We go beyond the ROI features and systematically study how other visual features that encode geometrical and visibility information can improve the network performance, where a simple yet effective four-way feature fusion scheme is adopted. Equipped with a dense 3D shape representation achieved by a~\textit{divide-and-conquer} strategy, we further design a multi-objective loss function to effectively improve network learning as validated by extensive experiments. This loss function considers geometric consistency using the projection of 66 semantic keypoints instead of masks which effectively reduces the ambiguity issue. It also incorporates a scene-aware term considering both inter-instance and instance-environment constraints.

In summary, our contributions are:
(1) A novel end-to-end network that can jointly reconstruct 3D pose and dense shape of vehicles, achieving state-of-the-art performance on the largest multi-task ApolloCar3D benchmark~\cite{song2019apollocar3d}.
(2) We propose an effective approach to extract and fuse diverse visual features, where systematic ablation study is shown to validate its effectiveness.
(3) GSNet reconstructs fine-grained 3D meshes (1352 vertices) by our \textit{divide-and-conquer} shape representation for vehicle instances rather than just 3D bounding boxes, wireframes~\cite{zia2013detailed} or retrieval~\cite{chabot2017deep,song2019apollocar3d}. 
(4) We design a new hybrid loss function to promote network performance, which considers both geometric consistency and scene constraints. This loss is made possible by the dense shape reconstruction, which in turn promotes the 6D pose estimation precision and sheds light on the benefit of jointly performing both tasks.

%------------------------------------------------------------------------
\section{Related Work}
\smallskip\noindent\textbf{Monocular 6DoF pose estimation.}
Traditionally, 6D object pose estimation is handled by creating correspondences between the object’s known 3D model and 2D pixel locations, followed by Perspective-n-Point (PnP) algorithm~\cite{rothganger20063d,wagner2008pose,peng2019pvnet}. 
For recent works,~\cite{cao2016real,hinterstoisser2011gradient} construct templates and calculate the similarity score to obtain the best matching position on the image. In~\cite{pavlakos20176,rothganger20063d,xiang2017posecnn}, 2D regional image features are extracted and matched with the features on 3D model to establish 2D-3D relation which thus require sufficient textures for matching.
A single-shot deep CNN is proposed in~\cite{tekin2018real} which regresses 6D object pose in one stage while in~\cite{sundermeyer2018implicit} a two-stage method is used: 1) SSD~\cite{liu2016ssd} for detecting bounding boxes and identities; 2) augmented autoencoder predicts object rotation using domain randomization~\cite{tobin2017domain}.
Most recently, Hu et al.~\cite{hu2019segmentation} introduces a segmentation-based method by combining local pose prediction from each visible part of the objects.
Comparing to the cases in self-driving scenarios, these methods~\cite{kehl2017ssd,xiang2017posecnn,rad2017bb8} are applied to indoor scenes with a small variance in translation especially along the longitudinal axis.
Although using keypoints information, our model does \textit{not} treat pose estimation as a PnP problem and is trained end-to-end.

\smallskip\noindent\textbf{Monocular 3D shape reconstruction.}
With the advent of large-scale shape datasets~\cite{chang2015shapenet} and the progress of data-driven approaches, 3D shape reconstruction from a single image based on convolutional neural networks is drawing increasing interests. Most of these approaches~\cite{richter2018matryoshka,wu20153d,zhu2017rethinking,yan2016perspective,sinha2017surfnet,riegler2017octnet,kong2017using,lin2019photometric} focus on general objects in the indoor scene or in the wild~\cite{kar2015category}, where single object is shot in a close distance and occupies the majority of image area. Different from them, GSNet is targeted for more complicated traffic environment with far more vehicle instances to reconstruct per image, where some of them are even under occlusion at a long distance~(over 50 meters away). 

\smallskip\noindent\textbf{Joint vehicle pose and shape reconstruction.}
3D traffic scene understanding from a single RGB image is drawing increasing interests in recent years. 
However, many of these approaches only predict object orientation with 3D bounding boxes~\cite{chen2017multi,liang2018deep,yang2018pixor,liu2019deep,brazil2019m3d,simonelli2019disentangling,xu2018multi}.
When it comes to 3D vehicle shape reconstruction, since the KITTI dataset~\cite{geiger2012we} labels cars using 3D bounding boxes with no detailed 3D shape annotation, existing works mainly use wireframes~\cite{zia2013detailed,li2017deep,wu2016single,Krishna_ICRA2017} or retrieve from CAD objects~\cite{chabot2017deep,song2019apollocar3d,mottaghi2015coarse,xiang2015data}.
In~\cite{zeeshan2014cars}, the authors utilize 3D wireframe vehicle models to jointly estimate multiple objects in a scene and find that more detailed representations of object shape are highly beneficial to 3D scene understanding.
DeepMANTA~\cite{chabot2017deep} adopts a coarse-to-fine refinement strategy to first regress 2D bounding box positions and generate 3D bounding boxes and finally obtain pose estimation results via 3D template fitting~\cite{lepetit2009epnp} by using the matched skeleton template to best fit its 2D image observations, which requires no image with 3D ground truth.
\textit{Most related to ours}, 3D-RCNN~\cite{kundu20183d} regresses 3D poses and deformable shape parameters in a single forward pass, but it uses coarse voxel shape representation and the proposed render-and-compare loss causes ambiguity during training. Direct-based~\cite{song2019apollocar3d} further augments 3D-RCNN by adding mask pooling and offset flow.
In contrast to these prior works, GSNet produces a more fine-grained 3D shape representation of vehicles by effective four-way feature fusion and~\textit{divide-and-conquer} shape reconstruction, which further inspires a geometrical scene aware loss to regularize network training with rich supervisory signals.
%------------------------------------------------------------------------

\section{Pose and Shape Representation}
\label{sec:representation} 
\smallskip\noindent\textbf{6DoF Pose.}
The 6DoF pose for each instance consists of the 3D translation $\mathbf{T}$ and 3D rotation $\mathbf{R}$. $\bR{T}$ is represented by the object center coordinate ${C}_{obj}=\{x,y,z\}$ in the camera coordinate system ${C}_{cam}$.
Rotation $\mathbf{R}$  defines the rotation Euler angles about ${X,Y,Z}$ axes of the object coordinate system ${C}_{obj}$.

\smallskip\noindent\textbf{Divide-and-Conquer Shape Representation.} 
We represent vehicle shape with dense mesh consisting of 1352 vertices and 2700 faces, which is much more fine-grained compared to the volume representation used in \cite{kundu20183d}. We start with the CAD meshes provided by the ApolloCar3D database~\cite{song2019apollocar3d}, which has different topology and vertex number for each car type. We convert them into the same topology with a fixed number of vertices by deforming a sphere using the SoftRas~\cite{liu2019soft} method.

\begin{figure}[!t]
	\centering
	\includegraphics[width=0.9\linewidth]{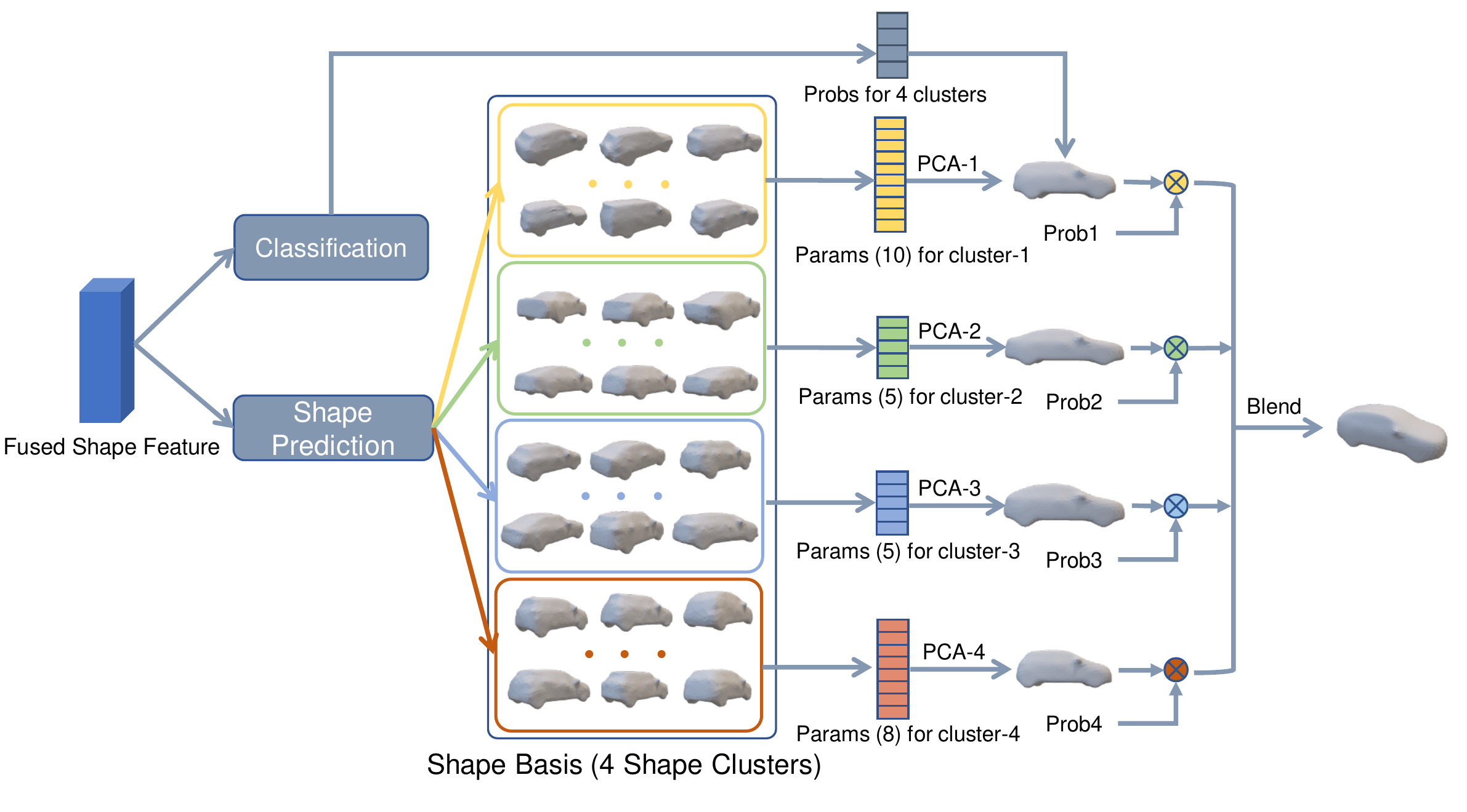}
	\caption{Illustration of our divide-and-conquer 3D shape reconstruction module, where we obtain four independent PCA models for each shape cluster. Instance shape reconstruction is achieved by reconstructing shape in each cluster and blend them with the respective classification probabilities. This strategy achieves lower shape reconstruction error compared to other methods as shown in Table~\ref{tab:compare_shape}.}
	\label{fig:example2}
\end{figure}

To ease the training of neural network for shape reconstruction, we reduce the shape representation dimension with principle component analysis (PCA)~\cite{prisacariu2011nonlinear}. However, applying PCA to all available meshes directly~\cite{engelmann2017samp,leotta2010vehicle,leotta2009predicting} is sub-optimal due to the large variation of car types and shapes. We thus adopt a \textit{divide-and-conquer} strategy as shown in Figure~\ref{fig:example2}. We first cluster a total of 79 CAD models into four subsets with K-Means algorithm utilizing the shape similarity between car meshes. For each subset, we separately learn a low dimensional shape basis with PCA.
Denote a subset of $k$ vehicle meshes as $M = \{m_1,m_2,...,m_k\}$, we use PCA to find $n\leq10$ dimensional shape basis, $\bar{\bR{S}} \in \mathbb{R}^{N\times n}$, where $N \gg n$.
During inference, the network classifies the input instance into the 4 clusters and predicts the principle component coefficient for each cluster. The final shape is blended from the four meshes weighted by the classification score. With this strategy, we achieve lower shape reconstruction error than directly applying PCA to all meshes or retrieval which is detailed in our ablation study.
\label{sec:pca}

\section{Network Architecture Design}
\begin{figure*}[!t]
	\centering
	\includegraphics[width=1.0\linewidth]{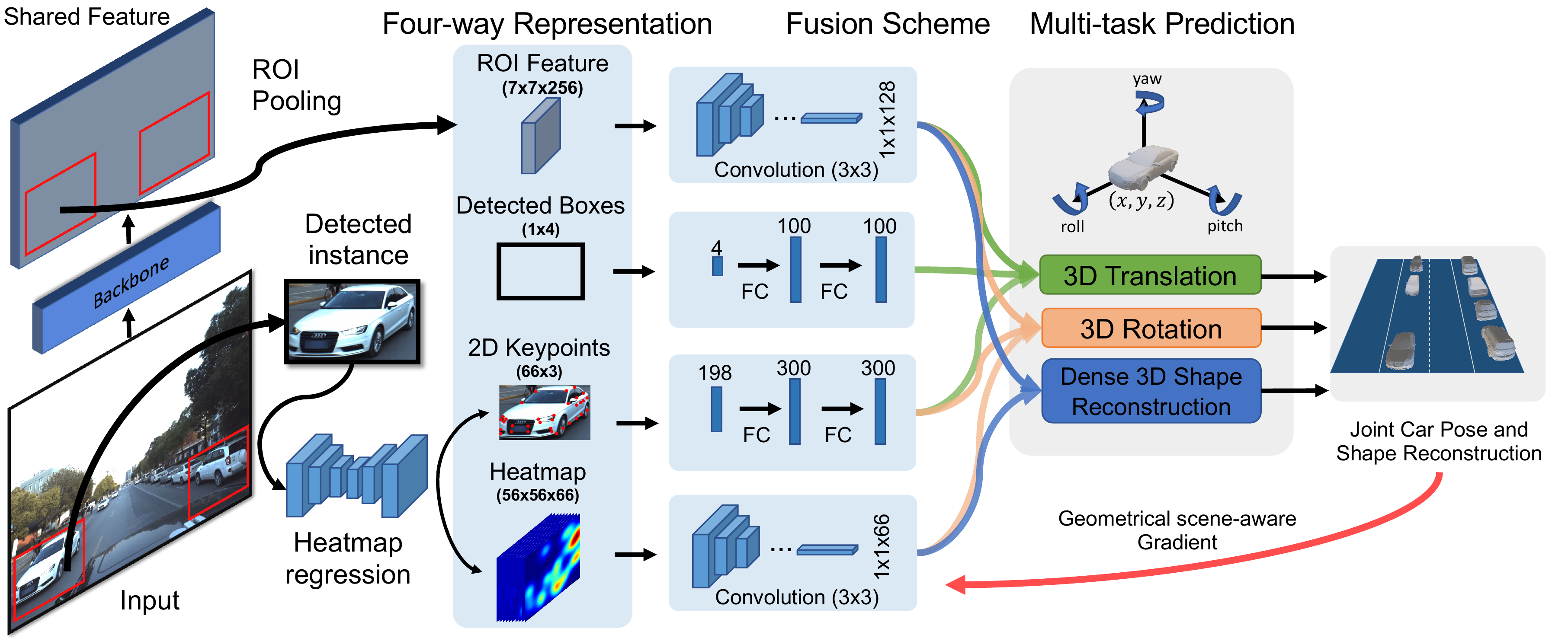}
	\caption{Overview of our GSNet for joint vehicle pose and shape reconstruction. We use region-based 2D object detector~\cite{he2017mask} and a built-in heatmap regression branch to obtain ROI features, detected boxes, keypoint coordinates (global locations in the whole image) and corresponding heatmap (local positions and visibility in sub-region). GSNet performs an effective fusion of four-way input representations and builds three parallel branches respectively for 3D translation, rotation and shape estimation. 3D shape reconstruction is detailed in Figure~\ref{fig:example2} and our hybrid loss function is illustrated in section~\ref{sec:constraint}.}
	\label{fig:model}
\end{figure*}

Figure \ref{fig:model} shows the overall architecture of our GSNet for joint car pose and shape reconstruction. We design and extract four types of features from a complex traffic scene, after which a fusion scheme is proposed to aggregate them. Finally, multi-task prediction is done in parallel to estimate 3D translation, rotation and shape via the intermediate fused representations.

\smallskip\noindent\textbf{Diverse Feature Extraction and Representation.}
Existing methods~\cite{kundu20183d,xiang2017posecnn} only use ROI features to regress 3D parameters, but we argue that using diverse features can better extract useful information in a complex traffic scene. Given an input image, we first use a region-based 2D object detector~\cite{he2017mask} to detect car instances and obtain its global location. Based on the bounding boxes, ROI pooling is used to extract appearance features for each instance. In a parallel branch, each detected instance is fed to a fully-convolutional sub-network to obtain 2D keypoint heatmaps and coordinates. The coordinates encode rich geometric information that can hardly be obtained with ROI features alone~\cite{zhao2017simple}, while the heatmaps encode part visibility to help the network discriminate occluded instances. 

Detected boxes are represented as 2D box center $(b_{x},b_{y})$, width $b_w$ and height $b_h$ in pixel space.
Camera intrinsic calibration matrix is $[f_x, 0, p_x;0, f_y, p_y; 0, 0, 1 ]$ where $f_x, f_y$ are focal lengths in pixel units and $(p_x, p_y)$ is the principal point at the image center.
We transform $b_{x},b_{y},b_w,b_h$ from pixel space to the corresponding coordinates $u_{x},u_{y},u_w,u_h$ in the world frame: %, the formula is defined as:
\begin{equation}
u_{x}=\frac{(b_{x}-p_x)z}{f_x},u_{y}=\frac{(b_{y}-p_y)z}{f_y},u_w=\frac{b_w}{f_x},u_h=\frac{b_h}{f_y},
\end{equation}
where $z$ is the fixed scale factor. For keypoint localization, we use the 66 semantic keypoints for cars defined in~\cite{song2019apollocar3d}. A 2D keypoint is represented as $\mathbf{p_k}=\{x_k,y_k,v_k\}$, where $\{x_k,y_k\}$ are the image coordinates and $v_k$ denotes visibility. In implementation, we adapt~\cite{he2017mask} pre-trained for human pose estimation on COCO to initialize the keypoint localization branch. For extracting ROI features, we use FPN~\cite{lin2017feature} as our backbone.

\smallskip\noindent\textbf{Fusion Scheme.}
We convert the extracted four-way inputs into 1D representation separately and decide which features to use for completing each task by prior knowledge. 
For global keypoint positions and detected boxes, we apply two fully-connected layers to convert them into higher level feature. 
For ROI feature maps and heatmaps, we adopt sequential convolutional operations with stride 2 to reduce their spatial size to $1\times1$ while keeping the channel number unchanged.

Instead of blindly using all features for prediction, we fuse different feature types that are most informative for each prediction branch.
The translation $\mathbf{T}$ mainly affects the object location and scale during the imaging process, thus we concatenate the ROI feature, 2D keypoint feature and box position feature for translation regression.
The rotation $\mathbf{R}$ determines the image appearance of the object given its 3D shape and texture, thus we utilize the fusion of ROI feature, heatmap feature and the keypoint feature as input. For estimating shape parameters $\mathbf{S}$, we aggregate the ROI and heatmap features.

\subsubsection{Multi-task Prediction}
\label{sec:output} 
We design three~\textit{parallel} estimation branches (translation, rotation and shape reconstruction) as shown in Figure~\ref{fig:model} since they are independent, where each branch directly regresses the targets with mutual benefits. Note that parts of input features such as ROI heatmap and keypoint positions are shared in different branches, which can be jointly optimized and is beneficial as shown by our experiments. In contrast to previous methods that predict translation or depth using a discretization policy~\cite{fu2018deep}, GSNet can directly regress the translation vector and achieve accurate result \textit{without any post processing} or further refinement. For the shape reconstruction branch, the network classify the input instance and estimates the low-dimensional parameters (less than 30) for four clusters as described in section~\ref{sec:representation}. 

\section{Geometrical and Scene-aware Supervision}
%\subsection{Geometric-Semantic-Aware Loss}
%\vspace{-1mm}
\label{sec:constraint} 
To provide GSNet with rich supervisory signals, we design a composite loss functions consisting of multiple terms. Apart from ordinary regression losses, it also strives for geometrical consistency and considers scene-level constraints in both inter- and intra-instance manners.
% which is beneficial to accurate translation estimation along longitudinal axis.

\begin{figure}[!t]
	\centering
	%\fbox{\rule{0pt}{2in} \rule{0.9\linewidth}{0pt}}
	\includegraphics[width=0.9\linewidth]{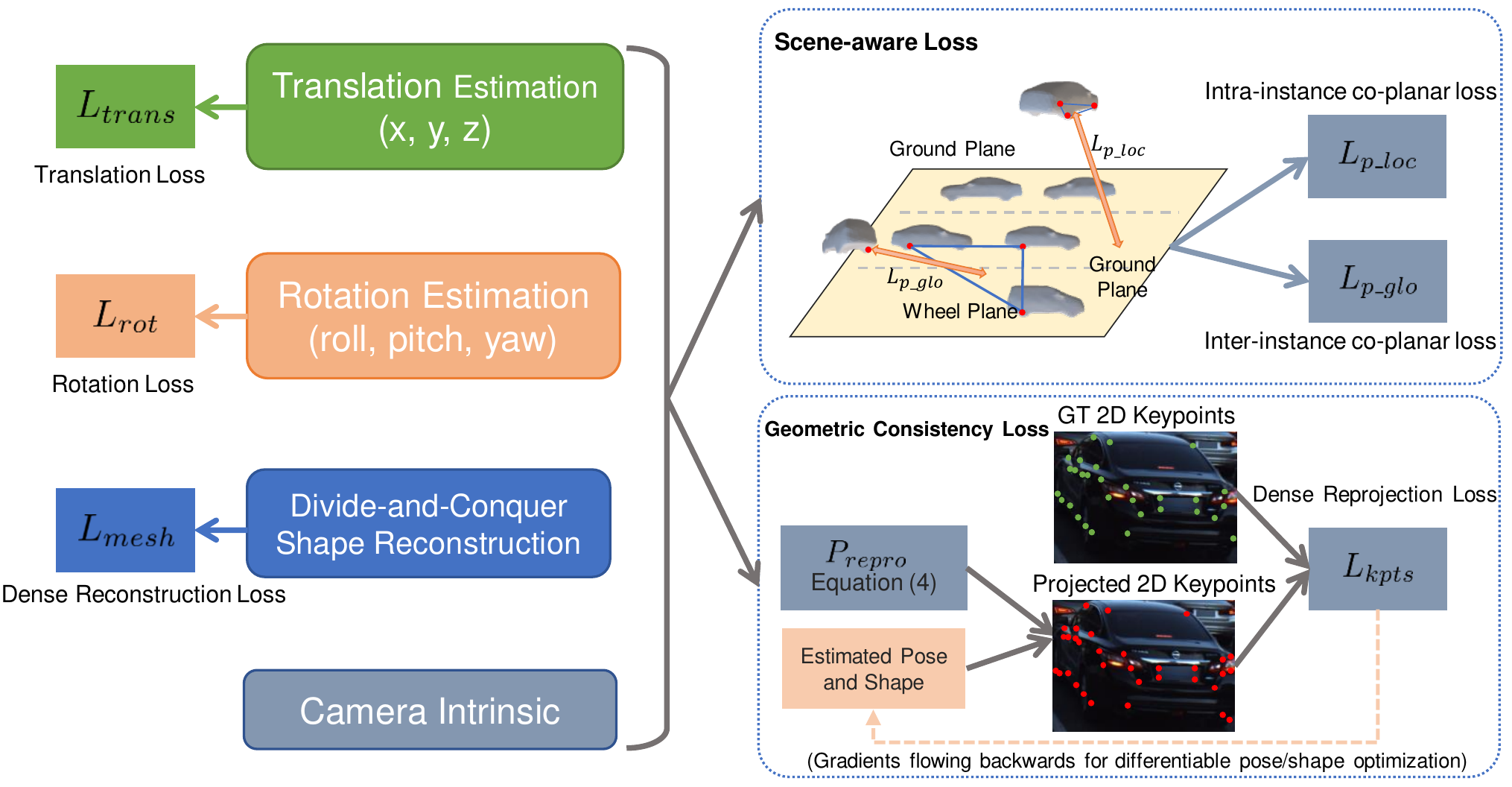}
	\caption{The hybrid loss function for optimizing the GSNet. The scene-aware loss consists of two parts,  $L_{p\_{glo}}$ for multiple car instances resting on common ground and $L_{p\_{loc}}$ for each single car at \textit{a fine-grained level}. For geometrical consistency, camera intrinsics are used to project the predicted 3D semantic vertices on a car mesh to image and compared with the 2D detections.} 
	\label{fig:example3}
	%\vspace{-0.25in}
\end{figure}

\smallskip\noindent\textbf{Achieve geometrical consistency by projecting keypoints.}
With the rich geometric details of the 3D vehicles as shown in Figure~\ref{fig:example3}, we exploit the 2D-3D keypoints correspondence using a pinhole camera model to provide extra supervision signal. For a 3D semantic keypoint $\bR{p}_k=(x_0,y_0,z_0)$ on the predicted mesh with translation $\bR{T}_{pred}$ and rotation $\bR{R}_{pred}$, the reprojection equation is:
\begin{equation}
P_{repro}= s\begin{bmatrix}
u_0\\ 
v_0\\ 
1
\end{bmatrix} =\bR{k}[\bR{R}_{pred}|\bR{T}_{pred}]\bR{p}_k,
\end{equation}
where $\bR{k}$ is the camera intrinsic matrix and $(u_0,v_0)$ is the projection point in pixels units. 
For the $i$th projected keypoint $\bR{p}_{i} = (u_i,v_i)$, the reprojection loss is
\begin{equation}
L_{kpt\_i} = \left \| \bR{p}_i-\bR{\bar{p}}_{i} \right \|_2^2,
\end{equation}
where $\bR{\bar{p}}_{i} = (\bar{u}_{i},\bar{v}_{i})$ is the corresponding image evidence given by our heatmap regression module.
The total loss $L_{kpts}$ for $n$ semantic keypoints in a car instance is
\begin{equation}
L_{kpts}= \sum_{i=1}^{n}{L_{kpt\_i}V_i},
\end{equation}
where $V_i$ is a boolean value indicating the visibility of $i$th keypoint in the image. This reprojection loss is \textit{differentiable} and can be easily incorporated in the end-to-end training process. The correspondence of 2D keypoints and 3D mesh vertices is needed to compute the loss and we determine it by ourselves. We project each 3D vertex on the ground truth mesh to image plane and find its nearest neighboring 2D points. The 66 3D vertices whose 2D projections have the most 2D annotated neighbors are selected as the corresponding 3D landmarks. We also provide an ablation experiment on the influence of keypoints number in the supplementary file.

\smallskip\noindent\textbf{Scene-aware Loss.}
Observe that most of cars rest on a common ground plane and the height of different instances is similar, thus the car centers are nearly co-planar. For each image, we locate mesh centers for four randomly-selected instances. Three of the centers define a plane $ax+by+cz+d=0$ and denote the remaining car center coordinate as $(x_1,y_1,z_1)$. As shown in Figure~\ref{fig:example3}, we introduce the \textit{inter-instance co-planar loss} $L_{p\_{glo}}$ for multiple cars  as:
\begin{equation}
L_{p\_{glo}} =\frac{|ax_1+by_1+cz_1+d|}{\sqrt{a^2+b^2+c^2}},
\label{eq:eq3}
\end{equation}

In addition, the centroids of the four wheels on a car should also lie in the same plane parallel to the ground. Thanks to the \textit{dense} 3D mesh reconstructed by our multi-task network, we can readily obtain these four 3D coordinates. We thus propose the~\textit{intra-instance co-planar loss} $L_{p\_{loc}}$ to supplements $L_{p\_{glo}}$. It is similar to Eq.~\ref{eq:eq3} but the three points are chosen on the same instance. 

\smallskip\noindent\textbf{Regression Losses.}
We use L2 loss $L_{mesh}$ to penalize inaccurate 3D shape reconstruction as:
\begin{equation}
L_{mesh} = \frac{\sum_{j=1}^{m}{\left \| \bR{M}_j-\bR{\bar{M}}_{j} \right \|_2^2}}{m},
\end{equation}
where $m$ is total number of vertices, $\bR{M}_j$ is the $j$th predicted vertex and $\bR{\bar{M}}_{j}$ is the ground truth vertex. For regression of 6DoF pose, we find that L1 loss performs better than L2 loss.The loss for translation regression is
\begin{equation}
L_{trans} = \left| \bR{T}_{pred} - \bR{T}_{gt}\right|,
\end{equation}
where $\bR{T}_{gt}$ and $\bR{T}_{pred}$ are ground-truth and predicted translation vector, respectively. 
For regressing rotation in Euler angles, we restrict the range around each axis $[-\pi,\pi]$. Since this is a unimodal task, we define the regression loss as
\begin{equation}
L_{rot} = \begin{cases}
\left |\bR{R}_{pred} - \bR{R}_{gt}  \right | & \text{ if } \left |\bR{R}_{pred} - \bR{R}_{gt}  \right | \leq  \pi,\\ 
2\pi - \left |\bR{R}_{pred} - \bR{R}_{gt}  \right | & \text{ if } \left |\bR{R}_{pred} - \bR{R}_{gt}  \right | > \pi,
\end{cases}
\end{equation}
where $\bR{R}_{pred}$ and $\bR{R}_{gt}$ are the predicted and ground truth rotation vector.

\smallskip\noindent\textbf{Sub-type Classification Loss.}
We also classify the car instance into 34 sub-types (sedan, minivan, SUV, etc.) and denote the classification loss as $L_{cls}$.

\smallskip\noindent\textbf{Final Objective Function.}
The final loss function $L$ for training our GSNet is defined as:
\begin{equation}
\begin{array}{l}
L = \lambda_{loc}L_{p\_{loc}} + \lambda_{glo}L_{p\_{glo}} + \lambda_{kpts}L_{kpts} \\
+ \lambda_{mesh}L_{mesh} + \lambda_{trans}L_{trans} +  \lambda_{rot}L_{rot} + \lambda_{cls}L_{cls}
\end{array}
\label{eq:eq10}
\end{equation}
where $\lambda$s balance the above loss components. As validated by our experiments in section~\ref{sec:ablation}, this hybrid loss function design significantly promotes the network's performance compared to using only regression losses alone.
%------------------------------------------------------------------------

\section{Experiments}
\label{sec:exp} 
\subsection{Datasets and Experimental Settings}

\smallskip\noindent\textbf{ApolloCar3D.}
We use the most recent and largest multi-task ApolloCar3D dataset~\cite{song2019apollocar3d} to train and evaluate GSNet. This dataset contains 5,277 high-resolution (2,710$\times$3384) images. We follow the official split where 4036 images are used for training, 200 for validation and the remaining 1041 for testing. Compared to KITTI~\cite{geiger2012we}, the instance count in ApolloCar3D is \textbf{20X} larger with far more cars per image (11.7 vs 4.8) where distant instances over 50 meters are also annotated. In addition,  ApolloCar3D provides 3D shape ground truth to evaluate shape reconstruction quantitatively, which is not available in KITTI.

\smallskip\noindent\textbf{Pascal3D+}
We also train and evaluate GSNet on Pascal3D+~\cite{xiang2014beyond} dataset using its car category. There are totally 6704 in-the-wild images with 1.19 cars per image on average. It also provides both dense 3D shape and 6D pose annotation.

\smallskip\noindent\textbf{Evaluation Metrics.}
We follow the evaluation metrics in~\cite{song2019apollocar3d}, which utilizes instance 3D average precision (A3DP)  with 10 thresholds (criteria from loose to strict) for \textbf{jointly} measuring translation, rotation and 3D car shape reconstruction accuracy. The results on the loose and strict criterion are respectively denoted as~\textit{c-l} and \textit{c-s}. During evaluation, Euclidean distance is used for 3D translation while arccos distance is used for 3D rotation. For 3D shape reconstruction, a predicted mesh is rendered into 100 views to compute IoU with the ground truth masks and the mean IoU is used. In addition to the absolute distance error, the relative error in translation is also evaluated to emphasize the model performance for nearby cars, which are more important for autonomous driving. We denote A3DP evaluated in relative and absolute version as~\textit{A3DP-Rel} and~\textit{A3DP-Abs} respectively.

\smallskip\noindent\textbf{Implementation Details.}
GSNet utilizes the Mask R-CNN~\cite{he2017mask} with ResNet-101 backbone pre-trained on the COCO 2017 dataset~\cite{lin2014microsoft} for object detection and extracting ROI features ($7\times7$). We discard detected objects with confidence score less than 0.3. The $\lambda_{loc}, \lambda_{glo},\lambda_{kpts},\lambda_{mesh}, \lambda_{trans}, \lambda_{rot},\lambda_{cls}$ in Eq.~\ref{eq:eq10} are set to 5.0, 5.0, 0.01, 10.0, 0.5, 1.0, 0.5 to balance the loss components.
During training, we use Adam optimizer~\cite{kingma2014adam} with initial learning rate 0.0025 and reduce it by half every 10 epochs for total 30 epochs. The 2D keypoint localization branch is trained separately where we use 4,036 training images containing 40,000 labeled vehicles with 2D keypoints and set threshold 0.1 for deciding keypoint visibility. When building the dense shape representation, there are respectively 9, 24, 14, 32 meshes in the four clusters.

\subsection{Ablation Study of Network Architecture and Loss Design}
\label{sec:ablation} 
We conduct three ablation experiments on ApolloCar3D validation set to validate our network design, loss functions and dense shape representation strategy.

\smallskip\noindent\textbf{Is extracting more features beneficial?}
We validate our four-way feature extraction fusion design by varying the number of used branches as:
1) Baseline: only using instance ROI features;
2) fusing transformed bounding box feature with the ROI feature;
3) combining predicted heatmap feature to the input;
4) further adding the 2D keypoint feature.
The quantitative comparison is shown in Table~\ref{table:ablation_input}. Compared to using ROI features alone, the injection of transformed detected boxes (center position, width and height) help provide geometric information, which help reduce translation error by 35.2\% while improves \textit{Rel-mAP} from 6.8 to 12.5 and \textit{Abs-mAP} from 7.0 to 11.4. 
The introduction of keypoint heatmaps is beneficial especially for rotation estimation. This extra visibility information for the 2D keypoints reduces rotation error by 25.0\% and further promoting \textit{Rel-mAP} from 12.5 to 13.7 and \textit{Abs-mAP} from 11.4 to 12.4. Finally, the 2D keypoint position branch complements the other three branches and improves model performance consistently for different evaluation metrics.

\smallskip\noindent\textbf{Effectiveness of the hybrid loss function.}
Here we fix our network architecture while varying the components of loss function to validate our loss design. 
The experiments are designed as follows:
1) Baseline: adopt four-way feature fusion architecture, but only train the network with regression and classification losses without shape reconstruction;
2) adding 3D shape reconstruction loss;
3) incorporating geometrical consistency loss;
4) adding scene-aware loss but only use the inter-instance version;
5) adding the intra-instance scene-aware component to complete the multi-task loss function.
As shown in Table~\ref{table:ablation_loss}, the reprojection consistency loss promotes 3D localization performance significantly, where the 3D translation error reduces over 10\% and \textit{Rel-mAp} increases from 15.1 to 17.6. The scene-aware loss brings obvious improvement compared to ignoring the traffic scene context, especially for the \textit{A3DP-Rel} strict criterion \textit{c-s} (increasing AP from 14.2 to 19.8). In addition, using both inter-instance and intra-instance loss components outperforms using inter-instance scene-aware loss alone. Compared to the baseline, our hybrid loss function significantly promotes the performance of \textit{Rel-mAp} and \textit{Abs-mAp} respectively to 20.2 and 18.9.

\smallskip\noindent\textbf{Is jointly performing both tasks helpful?} 
We argue that jointly performing dense shape reconstruction can in turn help 6D pose estimation. Without the introduction of the dense shape reconstruction task, we do not have access to the reconstruction loss C1 as well as the geometrical and scene-aware losses (C2, C3 and C4). Note that C1-C4 significantly improves estimation accuracy for translation and rotation. 
\begin{table*}[!t]
	\centering
	\tabcolsep=0.15cm
	\renewcommand{\arraystretch}{0.9}
	\caption{Ablation study for GSNet on four-way feature fusion, which shows the relevant contribution of each representation with only regression losses. Performance is evaluated in terms of A3DP (\textit{jointly} measuring translation, rotation and 3D car shape reconstruction accuracy), where \textit{c-l} indicates results on loose criterion and \textit{c-s} indicates strict criterion. GSNet exhibits a significant improvement compared to the baseline (with only ROI features),  which promotes \textit{A3DP-Rel} item \textit{c-s} from 3.2 to 10.5. T and R in \textit{6DoF Error} respectively represent 3D translation and rotation.
	}
	{   \footnotesize
		\resizebox{0.95\columnwidth}{!}{
			\begin{tabular}{ c | c | c | c | c | c | c | c | c | c | c | c }
				\toprule
				\multicolumn{4}{c|}{2D Input Representation} &
				\multicolumn{3}{c|}{A3DP-Rel} &
				\multicolumn{3}{c|}{A3DP-Abs} &
				\multicolumn{2}{c}{6DoF Error} \\
				\cline{1-12} 
				ROI & boxes & heatmap & kpts & mean & c-l & c-s & mean & c-l & c-s & T & R \\
				\hline
				\checkmark &  &  &  & 6.8 & 20.1 & 3.2 & 7.0 & 17.7  & 5.1 & 2.41 & 0.33 \\
				\checkmark & \checkmark  &  &  & 12.5$_{\uparrow5.7}$ & 30.1$_{\uparrow10.4}$ & 8.9$_{\uparrow5.7}$ & 11.4$_{\uparrow4.4}$ & 26.6$_{\uparrow8.9}$ & 8.8$_{\uparrow3.7}$ & 1.56$_{\downarrow0.85}$ & 0.32$_{\downarrow0.01}$ \\
				\checkmark & \checkmark & \checkmark &  & 13.7$_{\uparrow6.9}$ & 32.5$_{\uparrow12.4}$ & 9.2$_{\uparrow6.0}$ & 12.4$_{\uparrow5.4}$ & 29.2$_{\uparrow11.5}$ & 9.2$_{\uparrow4.1}$ & 1.53$_{\downarrow0.88}$ & 0.24$_{\downarrow0.09}$ \\
				\checkmark & \checkmark  & \checkmark & \checkmark &  \textbf{14.1}$_{\uparrow7.3}$ & \textbf{32.9}$_{\uparrow12.8}$ & \textbf{10.5}$_{\uparrow7.3}$ & \textbf{12.8}$_{\uparrow5.8}$ & \textbf{29.3}$_{\uparrow11.6}$  & \textbf{9.9}$_{\uparrow4.8}$ & \textbf{1.50}$_{\downarrow0.91}$ & \textbf{0.24}$_{\downarrow0.09}$ \\
				\bottomrule
			\end{tabular}
	}}
	\label{table:ablation_input}
\end{table*}

\begin{table*}[!t]
	\centering
	\tabcolsep=0.15cm
	\renewcommand{\arraystretch}{0.9}
	\caption{Ablation study for GSNet using different loss components of the hybrid loss function, which shows the relevant contribution of each component. C0, C1, C2, C3, C4 respectively denote pose regression loss, 3D shape reconstruction loss, geometrical consistency loss, inter-instance scene-aware loss and intra-instance scene-aware loss. GSNet exhibits a significant improvement compared to the baseline (with only regression losses), especially in estimating the surrounding car instances as shown by \textit{A3DP-Rel} (item \textit{c-s} has been significantly boosted from 10.5 to 19.8).
	}
	%\vspace{-0.1in}
	{   \footnotesize
		\resizebox{0.95\columnwidth}{!}{
			\begin{tabular}{ c | c | c | c | c | c | c | c | c | c | c | c | c}
				\toprule
				\multicolumn{5}{c|}{Loss Components} &
				\multicolumn{3}{c|}{A3DP-Rel} &
				\multicolumn{3}{c|}{A3DP-Abs} &
				\multicolumn{2}{c}{6DoF Error} \\
				\cline{1-13} 
				C0 & C1 & C2 & C3 & C4 & mean & c-l & c-s & mean & c-l & c-s & T & R \\
				\hline
				\checkmark &  &  &  & & 14.1 & 32.9 & 10.5 & 12.8 & 29.3  & 9.9 & 1.50 & 0.24 \\
				\checkmark & \checkmark  & &  & &  15.1$_{\uparrow1.0}$ & 34.8$_{\uparrow1.9}$ & 11.3$_{\uparrow0.8}$ &
				15.0$_{\uparrow2.2}$ & 32.0$_{\uparrow2.7}$  & 13.0$_{\uparrow3.1}$ & 1.44$_{\downarrow0.06}$ & 0.23$_{\downarrow0.01}$ \\
				\checkmark & \checkmark & \checkmark & &  &  17.6$_{\uparrow3.5}$ & 37.3$_{\uparrow4.4}$ & 14.2$_{\uparrow3.7}$ &
				16.7$_{\uparrow3.9}$ & 34.1$_{\uparrow4.8}$  & 15.4$_{\uparrow5.5}$ & 1.30$_{\downarrow0.20}$ & 0.20$_{\downarrow0.04}$ \\
				\checkmark & \checkmark & \checkmark & \checkmark &  &  18.8$_{\uparrow4.7}$ & 39.0$_{\uparrow6.1}$ & 16.3$_{\uparrow5.8}$ &
				17.6$_{\uparrow4.8}$ & 35.3$_{\uparrow6.0}$  & 16.7$_{\uparrow6.8}$ & 1.27$_{\downarrow0.23}$ & 0.20$_{\downarrow0.04}$ \\
				\checkmark & \checkmark & \checkmark & \checkmark & \checkmark & \textbf{20.2$_{\uparrow\textbf{6.1}}$ } & \textbf{40.5$_{\uparrow\textbf{7.6}}$ } & \textbf{19.8$_{\uparrow\textbf{9.3}}$ } & \textbf{18.9$_{\uparrow\textbf{6.1}}$ } & \textbf{37.4$_{\uparrow\textbf{8.1}}$ }  & \textbf{18.3$_{\uparrow\textbf{8.4}}$ } & \textbf{1.23$_{\downarrow\textbf{0.27}}$} & \textbf{0.18$_{\downarrow\textbf{0.06}}$} \\
				\bottomrule
			\end{tabular}
	}}
	\label{table:ablation_loss}
\end{table*}

\smallskip\noindent\textbf{Effectiveness of the divide-and-conquer strategy.}
Table~\ref{tab:compare_shape} compares model performance using different shape representations: retrieval, single PCA shape-space model and our divide-and-conquer strategy detailed in section~\ref{sec:representation}. Observe that our divide-and-conquer strategy not only reduces shape reconstruction error for around 10\%, but also boosts the overall performance for traffic instance understanding.  Also, we present shape reconstruction error distribution across different vehicle categories in our supplementary file.

\subsection{Comparison with state-of-the-art methods}
\label{sec:performance}

\subsubsection{Quantitative Comparison on ApolloCar3D}
\label{sec:quantitative_analysis}
We compare GSNet with state-of-the-art approaches that jointly reconstruct vehicle pose and shape on ApolloCar3D dataset as shown in Table~\ref{table:compare_other}. 
The most recent \textit{regression-based} approaches are:
1) 3D-RCNN~\cite{kundu20183d}, which regress 3D instances from ROI features and add geometrical consistency by designing a differentiable render-and-compare mask loss;
2) Direct-based method in ~\cite{song2019apollocar3d}, which improves 3D-RCNN by adding mask pooling and offset flow.
We can see that our GSNet achieves superior results among the existing \textit{regression-based} methods across the evaluation metrics while being fast, nearly doubling the mAP performance of 3D-RCNN in \textit{A3DP-Rel} entry. Compared to the~\textit{fitting-based} pose estimation methods using Epnp~\cite{lepetit2009epnp}, which fit 3D template car model to 2D image observations in a time-consuming optimization process, GSNet performs comparably in \textit{A3DP-Rel} and \textit{A3DP-Abs} metrics with a high-resolution shape reconstruction output not constrained by the existing CAD templates. Note that ~\textit{fitting-based} methods consume long time and thus are not feasible for time-critical applications. Also note that \textit{A3DP-Rel} is important since nearby cars are more relevant for self-driving car to make motion planning, where GSNet improves \textit{c-l} AP performance by 15.75 compared to Kpts-based~\cite{song2019apollocar3d}. 

\begin{table*}[!t]
	\centering
	\tabcolsep=0.15cm
	\renewcommand{\arraystretch}{0.9}
	\caption{Performance comparison with state-of-the-art 3D joint vehicle pose and shape reconstruction algorithms on ApolloCar3D dataset. Times is the average inference time for processing each image. GSNet achieves significantly better performance than state-of-the-art regression-based approaches (using a deep network to directly estimate the pose/shape from pixels) with both high precision and fast speed where inference time is~\textit{critical} in autonomous driving. * denotes fitting-based methods, which fits a 3D template car model to best match its 2D image observations (requires no image with 3D ground truth) and is time-consuming. } 
	{
		\resizebox{0.95\columnwidth}{!}{
			\begin{tabular}{ l | c | c | c | c | c | c | c | c | c | c}
				\toprule
				\multirow{2}{*}{Model} &
				\multirow{2}{*}{Shape Reconstruction} &
				\multirow{2}{*}{Regression-based} &
				\multicolumn{3}{c|}{A3DP-Rel} &
				\multicolumn{3}{c|}{A3DP-Abs} &
				\multirow{2}{*}{Times} &
				\multirow{2}{*}{Time-efficient} \\
				\cline{4-9} 
				& & & mean & c-l & c-s & mean & c-l & c-s & \\
				\hline
				DeepMANTA (CVPR'17)~\cite{chabot2017deep}$^*$ & retrieval & \xmark & 16.04 & 23.76 & 19.80 &20.10 & 30.69  & 23.76 & 3.38s & \xmark \\
				Kpts-based (CVPR'19)~\cite{song2019apollocar3d}$^*$  & retrieval & \xmark & 16.53 & 24.75 & 19.80 & 20.40 & 31.68  & 24.75 & 8.5s & \xmark \\
				\hline
				\hline
				3D-RCNN (CVPR'18)~\cite{kundu20183d} & TSDF volume~\cite{curless1996volumetric} & \cmark  & 10.79 & 17.82 & 11.88 &16.44 & 29.70  & \textbf{19.80} & 0.29s & \cmark \\
				Direct-based (CVPR'19)~\cite{song2019apollocar3d} & retrieval & \cmark & 11.49 & 17.82 & 11.88 &15.15 & 28.71  & 17.82 & 0.34s & \cmark \\
				\midrule
				Ours: GSNet & Detailed deformable mesh & \cmark & \textbf{20.21}  & \textbf{40.50} & \textbf{19.85} & \textbf{18.91} & \textbf{37.42} & 18.36 & 0.45s & \cmark \\
				\bottomrule
			\end{tabular}
	}}
	\label{table:compare_other}
\end{table*}

\begin{table}[!t]
	\begin{minipage}[t]{0.48\linewidth}
		\caption{Results comparison between GSNet adopting retrieval, single PCA model and our divide-and-conquer shape module on ApolloCar3D validation set.}
		\centering
		\resizebox{1.0\linewidth}{!}{
			\begin{tabular}{ c | c | c }
				\toprule
				Shape-space Model & Shape Reconstruction Error & Rel-mAP\\
				\midrule
				Retrieval & 92.46 & 17.6\\
				Single PCA & 88.68 & 18.7 \\
				\midrule
				Divide-and-Conquer Shape Module  & \textbf{81.33} & \textbf{20.2} \\
				\bottomrule
			\end{tabular}
		}
		\label{tab:compare_shape}
	\end{minipage}
	\label{tab:test}
	\hfill
	\begin{minipage}[t]{0.48\linewidth}
		\caption{Results on viewpoint estimation with annotated boxes on Pascal3D+~\cite{xiang2014beyond} for \textit{Car}, where GSNet gets highest accuracy and lowest angular error.}
		\centering
		\resizebox{0.8\linewidth}{!}{
			\begin{tabular}{ l | c | c }
				\toprule
				Model & $Acc_{\pi/6} \uparrow$ & $MedErr$ $\downarrow$\\
				\midrule
				RenderForCNN~\cite{su2015render} & 0.88 & 6.0$^{\circ}$ \\
				Deep3DBox~\cite{mousavian20173d}  & 0.90 & 5.8$^{\circ}$ \\
				3D-RCNN~\cite{kundu20183d} & 0.96 & 3.0$^{\circ}$ \\
				\midrule
				Ours: GSNet & \textbf{0.98} & \textbf{2.4}$^{\circ}$ \\
				\bottomrule
			\end{tabular}
		}
		\label{tab:compare_pascal}
	\end{minipage}
\end{table}

\begin{figure*}[!t]
	\centering
	\includegraphics[width=0.97\linewidth]{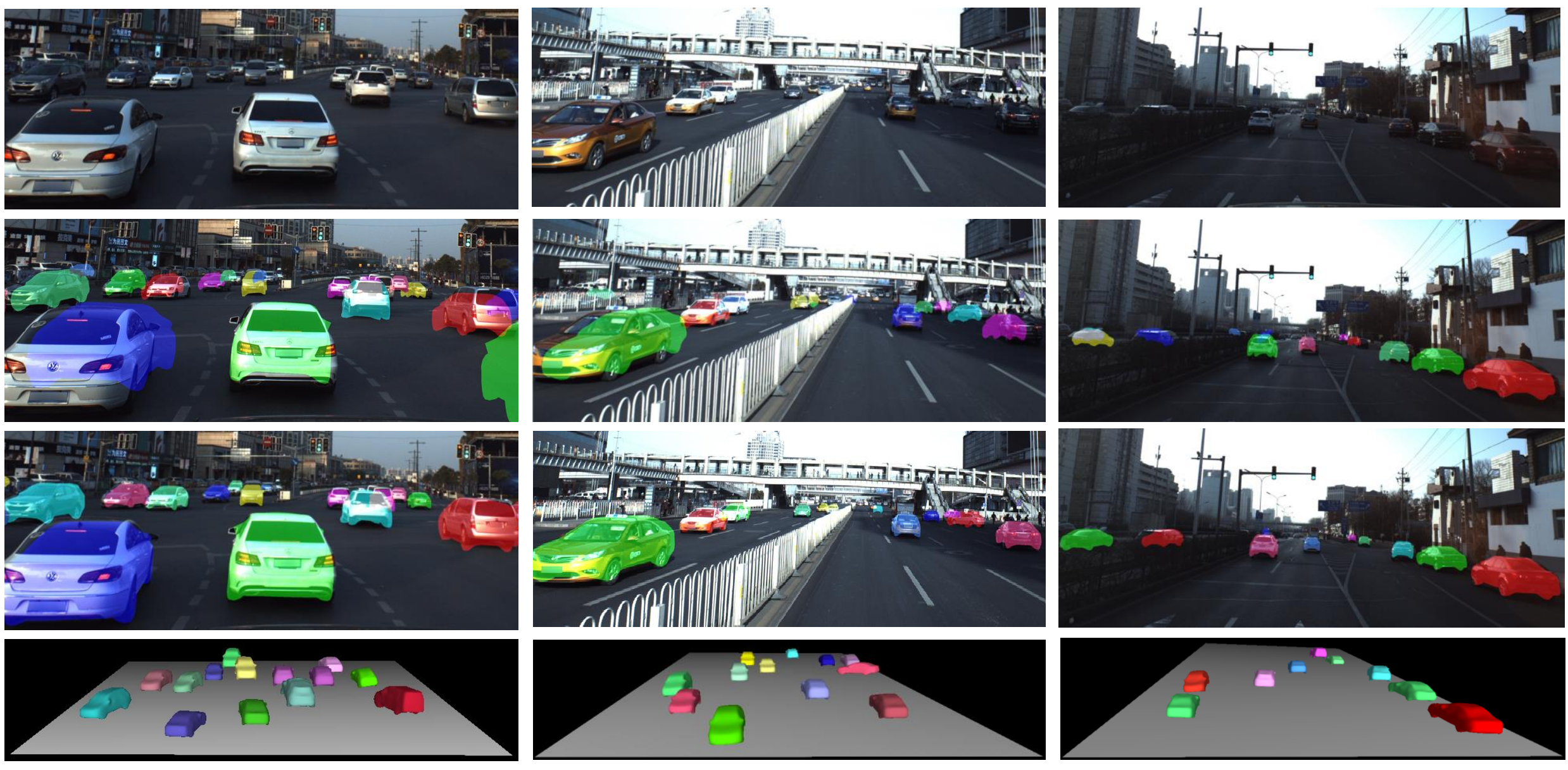}
	\caption{Qualitative comparison on the ApolloCar3D test set of different approaches by rendering 3D mesh output projected onto the input 2D image. The first row are the input images, the second row is the result of Direct-based~\cite{song2019apollocar3d} and the third row is predicted by our GSNet. The bottom row shows the reconstructed meshes in 3D space. Corresponding car instances are depicted in the same color.}
	\label{fig:qualitative_compare1}
\end{figure*}

\begin{figure*}[!t]
	\centering
	\includegraphics[width=0.97\linewidth]{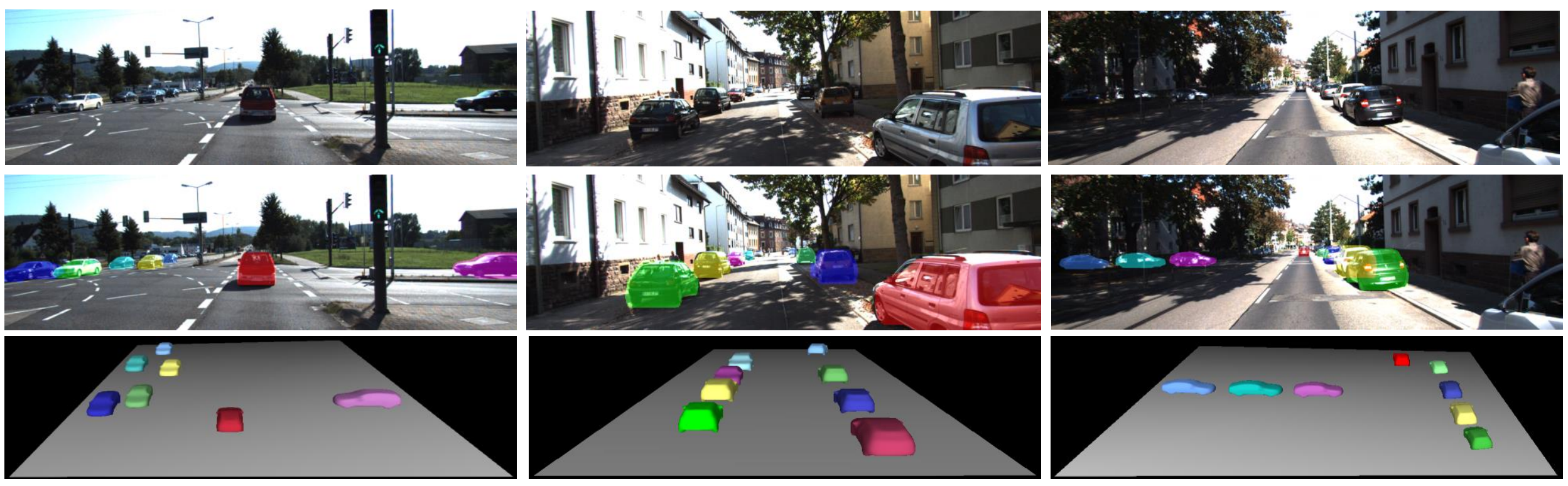}
	\caption{Cross-dataset generalization of GSNet on KITTI~\cite{geiger2012we} dataset. The first row are the input images and the second row are our reconstructed 3D car meshes projected onto the original image. Additional results are shown in our supplementary material.}
	\label{fig:qualitative_compare2}
\end{figure*}

\subsubsection{Quantitative Comparison on Pascal3D+}
To further validate our network, we evaluate GSNet on the Pascal3D+~\cite{xiang2014beyond} dataset using its car category. We follow the setting in~\cite{kundu20183d,mousavian20173d} to evaluate the viewpoint and use $Acc_{\pi/6}$ and $MedErr$ adopted in~\cite{mousavian20173d,tulsiani2015viewpoints} to report results in Table~\ref{tab:compare_pascal}, where the median angular error improves by 20\% from 3.0$^{\circ}$ to 2.4$^{\circ}$ compared to 3D-RCNN.

\subsubsection{Qualitative Analysis}
\label{sec:qualitative_analysis}
Figure~\ref{fig:qualitative_compare1} shows qualitative comparisons with other direct~\textit{regression-based} methods for joint vehicle pose and shape reconstruction. Compared with Direct-based~\cite{song2019apollocar3d}, our GSNet produces more accurate 6DoF pose estimation and 3D shape reconstruction from monocular images due to the effective four-way feature fusion, the hybrid loss which considers both geometrical consistency and scene-level constraints and our divide-and-conquer shape reconstruction. Although directly regressing depth based on monocular images is considered as an ill-posed problem, our GSNet achieves high 3D estimation accuracy (our projected masks of car meshes on input images show an almost perfect match), particularly for instances in close proximity to the self-driving vehicle.
The last column of the figure shows that the estimation of GSNet is still robust even in a relatively dark environment where the two left cars are heavily occluded. The last row visualizes the predicted 3D vehicle instances.

Figure~\ref{fig:qualitative_compare2} shows additional qualitative results on applying GSNet on KITTI~\cite{geiger2012we}. Despite that GSNet is not trained on KITTI, the generalization ability of our model is validated as can be seen from the accurate 6D pose estimation and shape reconstruction of unseen vehicles. More results (including 3 temporally preceding frames of KITTI) are available in our supplementary material.

\section{Conclusion}
We present an end-to-end multi-task network GSNet, which jointly reconstructs 6DoF pose and 3D shape of vehicles from single urban street view.
Compared to previous regression-based methods, GSNet not only explores more potential feature sources and uses an effective fusion scheme to supplement ROI features, but also provides richer supervisory signals from both geometric and scene-level perspectives.
Vehicle pose estimation and shape reconstruction are tightly integrated in our system and benefit from each other, where 3D reconstruction delivers geometric scene context and greatly helps improve pose estimation precision.
Extensive experiments conducted on ApolloCar3D and Pascal3D+ have demonstrated our state-of-the-art performance and validated the effectiveness of GSNet with both high accuracy and fast speed.
\subsubsection{Acknowledgement:}This research is supported in part by the Research Grant Council of the Hong Kong SAR under grant no. 1620818.

\clearpage
% ---- Bibliography ----
%
% BibTeX users should specify bibliography style 'splncs04'.
% References will then be sorted and formatted in the correct style.
%
\bibliographystyle{splncs04}
\bibliography{egbib}

\begin{thebibliography}{10}
\providecommand{\url}[1]{\texttt{#1}}
\providecommand{\urlprefix}{URL }
\providecommand{\doi}[1]{https://doi.org/#1}

\bibitem{brazil2019m3d}
Brazil, G., Liu, X.: M3d-rpn: Monocular 3d region proposal network for object
  detection. In: ICCV (2019)

\bibitem{cao2016real}
Cao, Z., Sheikh, Y., Banerjee, N.K.: Real-time scalable 6dof pose estimation
  for textureless objects. In: 2016 IEEE International conference on Robotics
  and Automation (ICRA) (2016)

\bibitem{chabot2017deep}
Chabot, F., Chaouch, M., Rabarisoa, J., Teuli{\`e}re, C., Chateau, T.: Deep
  manta: A coarse-to-fine many-task network for joint 2d and 3d vehicle
  analysis from monocular image. In: CVPR (2017)

\bibitem{chang2015shapenet}
Chang, A.X., Funkhouser, T., Guibas, L., Hanrahan, P., Huang, Q., Li, Z.,
  Savarese, S., Savva, M., Song, S., Su, H., et~al.: Shapenet: An
  information-rich 3d model repository. arXiv preprint arXiv:1512.03012  (2015)

\bibitem{chen2016monocular}
Chen, X., Kundu, K., Zhang, Z., Ma, H., Fidler, S., Urtasun, R.: Monocular 3d
  object detection for autonomous driving. In: CVPR (2016)

\bibitem{chen2017multi}
Chen, X., Ma, H., Wan, J., Li, B., Xia, T.: Multi-view 3d object detection
  network for autonomous driving. In: CVPR (2017)

\bibitem{cordts2016cityscapes}
Cordts, M., Omran, M., Ramos, S., Rehfeld, T., Enzweiler, M., Benenson, R.,
  Franke, U., Roth, S., Schiele, B.: The cityscapes dataset for semantic urban
  scene understanding. In: CVPR (2016)

\bibitem{curless1996volumetric}
Curless, B., Levoy, M.: A volumetric method for building complex models from
  range images. In: SIGGRAPH (1996)

\bibitem{engelmann2017samp}
Engelmann, F., St{\"u}ckler, J., Leibe, B.: Samp: shape and motion priors for
  4d vehicle reconstruction. In: WACV (2017)

\bibitem{fu2018deep}
Fu, H., Gong, M., Wang, C., Batmanghelich, K., Tao, D.: Deep ordinal regression
  network for monocular depth estimation. In: CVPR (2018)

\bibitem{geiger2012we}
Geiger, A., Lenz, P., Urtasun, R.: Are we ready for autonomous driving? the
  kitti vision benchmark suite. In: CVPR (2012)

\bibitem{he2017mask}
He, K., Gkioxari, G., Doll{\'a}r, P., Girshick, R.: Mask r-cnn. In: ICCV (2017)

\bibitem{hinterstoisser2011gradient}
Hinterstoisser, S., Cagniart, C., Ilic, S., Sturm, P., Navab, N., Fua, P.,
  Lepetit, V.: Gradient response maps for real-time detection of textureless
  objects. TPAMI  \textbf{34}(5),  876--888 (2011)

\bibitem{hu2019segmentation}
Hu, Y., Hugonot, J., Fua, P., Salzmann, M.: Segmentation-driven 6d object pose
  estimation. In: CVPR (2019)

\bibitem{kar2015category}
Kar, A., Tulsiani, S., Carreira, J., Malik, J.: Category-specific object
  reconstruction from a single image. In: CVPR (2015)

\bibitem{kehl2017ssd}
Kehl, W., Manhardt, F., Tombari, F., Ilic, S., Navab, N.: Ssd-6d: Making
  rgb-based 3d detection and 6d pose estimation great again. In: ICCV (2017)

\bibitem{kingma2014adam}
Kingma, D.P., Ba, J.: Adam: A method for stochastic optimization. In: ICLR
  (2015)

\bibitem{kolotouros2019learning}
Kolotouros, N., Pavlakos, G., Black, M.J., Daniilidis, K.: Learning to
  reconstruct 3d human pose and shape via model-fitting in the loop. In: ICCV
  (2019)

\bibitem{kong2017using}
Kong, C., Lin, C.H., Lucey, S.: Using locally corresponding cad models for
  dense 3d reconstructions from a single image. In: CVPR (2017)

\bibitem{Krishna_ICRA2017}
Krishna~Murthy, J., Sai~Krishna, G., Chhaya, F., Madhava~Krishna, K.:
  Reconstructing vehicles from a single image: Shape priors for road scene
  understanding. In: 2017 IEEE International Conference on Robotics and
  Automation (ICRA) (2017)

\bibitem{ku2019monocular}
Ku, J., Pon, A.D., Waslander, S.L.: Monocular 3d object detection leveraging
  accurate proposals and shape reconstruction. In: CVPR (2019)

\bibitem{kundu20183d}
Kundu, A., Li, Y., Rehg, J.M.: 3d-rcnn: Instance-level 3d object reconstruction
  via render-and-compare. In: CVPR (2018)

\bibitem{leotta2009predicting}
Leotta, M.J., Mundy, J.L.: Predicting high resolution image edges with a
  generic, adaptive, 3-d vehicle model. In: CVPR (2009)

\bibitem{leotta2010vehicle}
Leotta, M.J., Mundy, J.L.: Vehicle surveillance with a generic, adaptive, 3d
  vehicle model. TPAMI  \textbf{33}(7),  1457--1469 (2010)

\bibitem{lepetit2009epnp}
Lepetit, V., Moreno-Noguer, F., Fua, P.: Epnp: An accurate o(n) solution to the
  pnp problem. IJCV  \textbf{81}(2), ~155 (2009)

\bibitem{li2017deep}
Li, C., Zeeshan~Zia, M., Tran, Q.H., Yu, X., Hager, G.D., Chandraker, M.: Deep
  supervision with shape concepts for occlusion-aware 3d object parsing. In:
  CVPR (2017)

\bibitem{li2019stereo}
Li, P., Chen, X., Shen, S.: Stereo r-cnn based 3d object detection for
  autonomous driving. In: CVPR (2019)

\bibitem{li2018stereo}
Li, P., Qin, T., Shen, S.: Stereo vision-based semantic 3d object and
  ego-motion tracking for autonomous driving. In: ECCV (2018)

\bibitem{liang2018deep}
Liang, M., Yang, B., Wang, S., Urtasun, R.: Deep continuous fusion for
  multi-sensor 3d object detection. In: ECCV (2018)

\bibitem{lin2019photometric}
Lin, C.H., Wang, O., Russell, B.C., Shechtman, E., Kim, V.G., Fisher, M.,
  Lucey, S.: Photometric mesh optimization for video-aligned 3d object
  reconstruction. In: CVPR (2019)

\bibitem{lin2017feature}
Lin, T.Y., Doll{\'a}r, P., Girshick, R., He, K., Hariharan, B., Belongie, S.:
  Feature pyramid networks for object detection. In: CVPR (2017)

\bibitem{lin2014microsoft}
Lin, T.Y., Maire, M., Belongie, S., Hays, J., Perona, P., Ramanan, D.,
  Doll{\'a}r, P., Zitnick, C.L.: Microsoft coco: Common objects in context. In:
  ECCV (2014)

\bibitem{liu2019deep}
Liu, L., Lu, J., Xu, C., Tian, Q., Zhou, J.: Deep fitting degree scoring
  network for monocular 3d object detection. In: CVPR (2019)

\bibitem{liu2019soft}
Liu, S., Li, T., Chen, W., Li, H.: Soft rasterizer: A differentiable renderer
  for image-based 3d reasoning. In: ICCV (2019)

\bibitem{liu2016ssd}
Liu, W., Anguelov, D., Erhan, D., Szegedy, C., Reed, S., Fu, C.Y., Berg, A.C.:
  Ssd: Single shot multibox detector. In: ECCV (2016)

\bibitem{mottaghi2015coarse}
Mottaghi, R., Xiang, Y., Savarese, S.: A coarse-to-fine model for 3d pose
  estimation and sub-category recognition. In: CVPR (2015)

\bibitem{mousavian20173d}
Mousavian, A., Anguelov, D., Flynn, J., Kosecka, J.: 3d bounding box estimation
  using deep learning and geometry. In: CVPR (2017)

\bibitem{pavlakos20176}
Pavlakos, G., Zhou, X., Chan, A., Derpanis, K.G., Daniilidis, K.: 6-dof object
  pose from semantic keypoints. In: 2017 IEEE International Conference on
  Robotics and Automation (ICRA) (2017)

\bibitem{peng2019pvnet}
Peng, S., Liu, Y., Huang, Q., Zhou, X., Bao, H.: Pvnet: Pixel-wise voting
  network for 6dof pose estimation. In: CVPR (2019)

\bibitem{pohlen2017full}
Pohlen, T., Hermans, A., Mathias, M., Leibe, B.: Full-resolution residual
  networks for semantic segmentation in street scenes. In: CVPR (2017)

\bibitem{prisacariu2011nonlinear}
Prisacariu, V.A., Reid, I.: Nonlinear shape manifolds as shape priors in level
  set segmentation and tracking. In: CVPR (2011)

\bibitem{rad2017bb8}
Rad, M., Lepetit, V.: Bb8: A scalable, accurate, robust to partial occlusion
  method for predicting the 3d poses of challenging objects without using
  depth. In: ICCV (2017)

\bibitem{richter2018matryoshka}
Richter, S.R., Roth, S.: Matryoshka networks: Predicting 3d geometry via nested
  shape layers. In: CVPR (2018)

\bibitem{riegler2017octnet}
Riegler, G., Osman~Ulusoy, A., Geiger, A.: Octnet: Learning deep 3d
  representations at high resolutions. In: CVPR (2017)

\bibitem{rothganger20063d}
Rothganger, F., Lazebnik, S., Schmid, C., Ponce, J.: 3d object modeling and
  recognition using local affine-invariant image descriptors and multi-view
  spatial constraints. IJCV  \textbf{66}(3),  231--259 (2006)

\bibitem{simonelli2019disentangling}
Simonelli, A., Bul{\`o}, S.R.R., Porzi, L., L{\'o}pez-Antequera, M.,
  Kontschieder, P.: Disentangling monocular 3d object detection. In: ICCV
  (2019)

\bibitem{sinha2017surfnet}
Sinha, A., Unmesh, A., Huang, Q., Ramani, K.: Surfnet: Generating 3d shape
  surfaces using deep residual networks. In: CVPR (2017)

\bibitem{song2019apollocar3d}
Song, X., Wang, P., Zhou, D., Zhu, R., Guan, C., Dai, Y., Su, H., Li, H., Yang,
  R.: Apollocar3d: A large 3d car instance understanding benchmark for
  autonomous driving. In: CVPR (2019)

\bibitem{su2015render}
Su, H., Qi, C.R., Li, Y., Guibas, L.J.: Render for cnn: Viewpoint estimation in
  images using cnns trained with rendered 3d model views. In: ICCV (2015)

\bibitem{sundermeyer2018implicit}
Sundermeyer, M., Marton, Z.C., Durner, M., Brucker, M., Triebel, R.: Implicit
  3d orientation learning for 6d object detection from rgb images. In: ECCV
  (2018)

\bibitem{tekin2018real}
Tekin, B., Sinha, S.N., Fua, P.: Real-time seamless single shot 6d object pose
  prediction. In: CVPR (2018)

\bibitem{tobin2017domain}
Tobin, J., Fong, R., Ray, A., Schneider, J., Zaremba, W., Abbeel, P.: Domain
  randomization for transferring deep neural networks from simulation to the
  real world. In: 2017 IEEE/RSJ International Conference on Intelligent Robots
  and Systems (IROS) (2017)

\bibitem{tulsiani2015viewpoints}
Tulsiani, S., Malik, J.: Viewpoints and keypoints. In: CVPR (2015)

\bibitem{wagner2008pose}
Wagner, D., Reitmayr, G., Mulloni, A., Drummond, T., Schmalstieg, D.: Pose
  tracking from natural features on mobile phones. In: IEEE/ACM International
  Symposium on Mixed and Augmented Reality (2008)

\bibitem{wu2016single}
Wu, J., Xue, T., Lim, J.J., Tian, Y., Tenenbaum, J.B., Torralba, A., Freeman,
  W.T.: Single image 3d interpreter network. In: ECCV (2016)

\bibitem{wu20153d}
Wu, Z., Song, S., Khosla, A., Yu, F., Zhang, L., Tang, X., Xiao, J.: 3d
  shapenets: A deep representation for volumetric shapes. In: CVPR (2015)

\bibitem{xiang2015data}
Xiang, Y., Choi, W., Lin, Y., Savarese, S.: Data-driven 3d voxel patterns for
  object category recognition. In: CVPR (2015)

\bibitem{xiang2014beyond}
Xiang, Y., Mottaghi, R., Savarese, S.: Beyond pascal: A benchmark for 3d object
  detection in the wild. In: WACV (2014)

\bibitem{xiang2017posecnn}
Xiang, Y., Schmidt, T., Narayanan, V., Fox, D.: Posecnn: A convolutional neural
  network for 6d object pose estimation in cluttered scenes. Robotics: Science
  and Systems (RSS)  (2018)

\bibitem{xu2018multi}
Xu, B., Chen, Z.: Multi-level fusion based 3d object detection from monocular
  images. In: CVPR (2018)

\bibitem{yan2016perspective}
Yan, X., Yang, J., Yumer, E., Guo, Y., Lee, H.: Perspective transformer nets:
  Learning single-view 3d object reconstruction without 3d supervision. In:
  NIPS (2016)

\bibitem{yang2018pixor}
Yang, B., Luo, W., Urtasun, R.: Pixor: Real-time 3d object detection from point
  clouds. In: CVPR (2018)

\bibitem{yang2019std}
Yang, Z., Sun, Y., Liu, S., Shen, X., Jia, J.: Std: Sparse-to-dense 3d object
  detector for point cloud. In: ICCV (2019)

\bibitem{zeeshan2014cars}
Zeeshan~Zia, M., Stark, M., Schindler, K.: Are cars just 3d boxes?-jointly
  estimating the 3d shape of multiple objects. In: CVPR (2014)

\bibitem{zhao2017simple}
Zhao, R., Wang, Y., Martinez, A.M.: A simple, fast and highly-accurate
  algorithm to recover 3d shape from 2d landmarks on a single image. TPAMI
  \textbf{40}(12),  3059--3066 (2017)

\bibitem{zhu2017rethinking}
Zhu, R., Kiani~Galoogahi, H., Wang, C., Lucey, S.: Rethinking reprojection:
  Closing the loop for pose-aware shape reconstruction from a single image. In:
  ICCV (2017)

\bibitem{zia2013detailed}
Zia, M.Z., Stark, M., Schiele, B., Schindler, K.: Detailed 3d representations
  for object recognition and modeling. TPAMI  \textbf{35}(11),  2608--2623
  (2013)

\end{thebibliography}
%\bibliography{eccv2020submission}
\end{document}